\documentclass[11pt,a4paper]{article}
\usepackage[hyperref]{emnlp2020}
\usepackage{times}
\usepackage{latexsym}
\usepackage{hyperref}
\usepackage{booktabs}
\usepackage{amssymb}
\usepackage{graphicx}
\usepackage{subfigure}
\usepackage{float}

\usepackage{multirow}
\usepackage{natbib}

\usepackage{microtype}

\aclfinalcopy 

\title{On Task-Level Dialogue Composition of Generative Transformer Model}

\author{Prasanna Parthasarathi \thanks{ \ \ This work was done when the author was an intern at Google Brain. \texttt{pparth2@cs.mcgill.ca}} \\
  McGill University / Mila \\
   \\\And
  Arvind Neelakantan \thanks{ \ \ This work was done when the author was a Research Scientist at Google Brain.} \\
 OpenAI \\\And
  Sharan Narang \\
 Google Brain \\}
\date{}

\begin{document}
\maketitle
\begin{abstract}
Task-oriented dialogue systems help users accomplish tasks such as booking a movie ticket and ordering food via conversation. Generative models parameterized by a deep neural network are widely used for next turn response generation in such systems. It is natural for users of the system to want to accomplish multiple tasks within the same conversation, but the ability of generative models to compose multiple tasks is not well studied. In this work, we begin by studying the effect of training human-human task-oriented dialogues  towards improving the ability to compose multiple tasks on Transformer generative models. To that end, we propose and explore two solutions: (1) creating synthetic multiple task dialogue data for training from human-human single task dialogue and (2) forcing the encoder representation to be invariant to single and multiple task dialogues using an auxiliary loss. The results from our experiments highlight the difficulty of even the sophisticated variant of transformer model in learning to compose multiple tasks from single task dialogues. 

\end{abstract}

\section{Introduction}
Recent years have seen a tremendous surge in the application of deep learning methods for dialogue in general \cite{conv-seq2seq,pipeline,multiwoz,deal} and task-oriented dialogue \cite{sclstm, fb, neural_assistant} specifically. Task-oriented dialogue systems help users accomplish tasks such as booking a movie ticket and ordering food via conversation. Generative models are a popular choice for next turn response generation in such systems \cite{pipeline,latent,copy-dialog}. These models are typically learned using  large amounts of dialogue data for every task \cite{multiwoz,taskmaster}. It is natural for users of the task-oriented dialogue system to want to accomplish multiple tasks within the same conversation, e.g. booking a movie ticket and ordering a taxi to the movie theater within the same conversation. The brute-force solution would require collecting dialogue data for every task combination which might be practically infeasible given the combinatorially many possibilities.      

While the ability of generative dialogue models to compose multiple tasks has not yet been studied in the literature, there has been some investigation on the compositionality skills of deep neural networks. \citet{lake} propose a suite of tasks to evaluate a method's compositionality skills and find that deep neural networks generalize to unseen compositions only in a limited way. \citet{kottur} analyze whether the language emerged when multiple generative models interact with each other is compositional and conclude that compositionality arises only with strong regularization.   

Motivated by the practical infeasibility of collecting data for combinatorially many task compositions, we focus on task-level compositionality of text response generation models. We begin by studying the effect of training data size of human-human multiple task dialogues on the performance of Transformer \cite{transformer} generative models. Next, we explore two solutions to improve task-level compositionality. First, we propose a data augmentation approach \cite{aug_1, aug_2,alex,aug,back} where we create synthetic multiple task dialogues for training from human-human single task dialogue; we add a portion of one dialogue as a prefix to another to simulate multiple task dialogues during training. As a second solution, we draw inspiration from the domain adaptation literature \cite{domain_1,domain_2,domain_nlp_1,domain_nlp,video,speech_domain} and encourage the model to learn domain invariant representations with an auxiliary loss to learn representations that are invariant to single and multiple task dialogues. 

We conduct our experiments on the Multiwoz dataset \cite{multiwoz}. The dataset contains both single and multiple task dialogues for training and evaluation. In Multiwoz, the tasks in multiple task dialogues are only the combinations of tasks in single task dialogues. This allows the dataset to be an appropriate benchmark for our experiments. 

To summarize, our key findings are:
\begin{itemize}
\item{We study task-level compositionality of text response generation models and find that they are heavily reliant on multiple task conversations at train time to do well on such conversations at test time.}
\item{We explore two novel unsupervised solutions to improve task-level compositionality: (1) creating synthetic multiple task dialogue data from human-human single task dialogue and (2) forcing the encoder representation to be  invariant to single and multiple task dialogues using an auxiliary loss.}
\item{Highlighting the difficulty of composing tasks in generative dialogues with experiments on the Multiwoz dataset, where both the methods combined result only in a 8.5\% BLEU \cite{papineni2002bleu} score improvement when zero-shot evaluated on multiple task dialogues.}
\end{itemize}

\section{Background}

Let $d_1, d_2, \ldots, d_M$ be the dialogues in the training set and every dialogue $d_m = ((u^1_m, a^1_m), (u^2_m, a^2_m), \ldots, (u^{n_m}_m, a^{n_m}_m)$ ($\forall m \in \{1,2,\ldots,M\}$) consists of $n_m$ turns each of user and assistant. Further each user and assistant turn consists of a sequence of word tokens. The individual dialogue could be either single task or multiple task depending on the number of tasks being accomplished in the dialogue. 

The response generation model is trained to generate each turn of the assistant response given the conversation history. The generative model learns a probability distribution given by $P(a^{i} \mid (u^1, a^1), \ldots, (u^{i-1}, a^{i-1}), u^i)$. We drop the symbol $m$ that denotes a particular training  example for simplicity.   The assistant turn $a^i$ consists of a sequence of word tokens, $a^{i} = (w_1^i, w_2^i, \ldots, w_{l^i}^i)$. The response generation model factorizes the joint distribution left-to-right given by, 

$P(a^{i} \mid x^i) = \prod\limits_{j=1}^{l^i} P(w_j \mid x^i, w_1^i, \ldots, w_{j-1}^i)$ \\
where $x^i=((u^1, a^1), \ldots, (u^{i-1}, a^{i-1}), u^i)$ refers to the conversation history till the $i^{th}$ turn.

We use a Transformer \cite{transformer} sequence-to-sequence model to parameterize the above distribution. Given a training set of dialogues, the parameters of the Transformer model are learned to optimize the conditional language modelling objective given by, 
\begin{equation}
\label{eq:lm}
    L_{LM} =  \sum_{m=1}^{M} \sum _{i=1}^{n_m} \log P(a^{i} \mid x^i, \Theta)
\end{equation}
where $\Theta$ refers to the parameters of the Transformer model. 

\section{Data Augmentation}

The first solution we explore for task compositionality generates synthetic multiple task dialogues for training from human-human single task dialogues \footnote{\href{https://github.com/ppartha03/Dialogue-Compositionality-of-Generative-Transformer}{Code repository}}. Here, we sample two dialogues from the training set, and add a portion of one dialogue as a prefix to another. While this procedure might not create dialogues of the quality equivalent to human-human multiple task dialogue, it is an unsupervised way to create approximate multiple task dialogues that the model could theoretically benefit from.  

Concretely, we randomly sample two single task dialogues $d_i$ and $d_j$ from the training set and create a noisy multiple task dialogue by adding a fraction of the dialogue $d_j$ as a prefix to dialogue $d_i$. The fraction of dialogue taken from dialogue $d_j$ is given by the hyperparameter $augment\_fraction$. The number of times dialogue $d_i$ is augmented by a randomly sampled dialogue is given by the hyper-parameter $augment\_fold$. 

We consider two strategies for sampling the dialogue $d_j$. In $Random\_Augment$, the dialogue is uniformly randomly sampled from the remainder of the training set.  A potential issue with the random strategy is that it might create spurious task combinations and the model might fit to this noise. Motivated by the spurious task combination phenomenon, we consider another sampling strategy $Targeted\_Augment$ where we create synthetic multiple task dialogues only for task combinations that exist in the development set. Here, $d_j$ is sampled from a set of dialogues whose task is compatible with the task of dialogue $d_i$. The Transformer model is now trained on the augmented training set using the objective function given in Equation \ref{eq:lm}.  The effect of the sampling strategy and the  hyperparameters on the model performance is discussed  in the experiments section (Section \ref{sec:experiments}).

\section{Domain Invariant Transformer}
\label{sec:trans_disc}

\begin{figure}
    \centering
    \includegraphics[width=0.7\columnwidth,height=8cm]{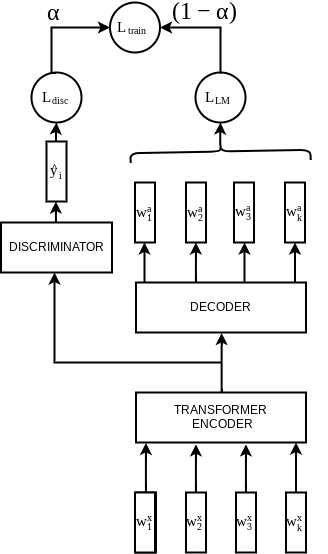}
    \caption{Domain Invariant Transformer Architecture.}
    \label{fig:trans_disc_diagram}
\end{figure}

We propose Domain Invariant Transformer model (Figure \ref{fig:trans_disc_diagram}) to maintain a domain invariant representation of the encoder by training the encoder representation for an auxiliary task. Here, the auxiliary task for the network is to predict the label ,$^i\hat{l}$, denoting the type of task (single or multi-task) in the encoded conversation history. The model takes as input the sequence of byte pair encoded tokens that are represented at the encoder hidden state as a set of attention weights from the multi-head multiple layer attention mechanism of transformer. The conditional language model (Equation \ref{eq:lm}) is learnt by a transformer decoder on top that attends over the encoder states.

The discriminator task network is trained with average pooling of the encoder summary over the attention heads ($h_j$)as shown in Equation \ref{averagepool}.

\begin{equation}
    ^ie^{s} = \sum_{j=1}^{k} \frac{\left(h_j\right)}{k}
    \label{averagepool}
\end{equation}

The average pooled encoder summary is passed as input to a two-layer feed forward discriminator. The discriminator network has a dropout \cite{srivastava14dropout} layer in-between the two fully connected layers ($f_1$ and $f_2$) (Equation \ref{discriminator}).   

\begin{equation}
    \hat{y}_i = f_2\left(f_1\left(^ie^{s}\right)\right) 
    \label{discriminator}
\end{equation}

The binary cross-entropy loss, $L_{disc}$, for the predicted label, $\hat{y}_i$, an input context \emph{i} is computed as in Equation \ref{disc_loss}.

\begin{equation}
    L_{disc} = - \left(y_i\log\left(\hat{y}_i\right) + \left(1 - y_i\right)\log\left(1 - \hat{y}_i\right)\right)
    \label{disc_loss}
\end{equation}

The Domain Invariant Transformer model optimizes a convex combination of the two losses as shown in Equation \ref{final_loss}. 

\begin{equation}
    L_{train} = \alpha * L_{disc} + \left(1-\alpha\right) * L_{LM}
\label{final_loss}
\end{equation}

The language model loss makes sure that the model learns to generate the next utterance while the discriminator loss makes sure the model is aware of the nature of task. To understand the effect of the auxiliary loss we experiment with different values for $\alpha$ (ref Appendix).




\section{Experiments}
\label{sec:experiments}

\subsection{Importance of multiple task dialogues}

We measure the importance of multiple task dialogue on the overall performance of transformer by training the model with varying amount of multiple task dialogues and keeping the task distribution between multiple and single domain dialogues almost similar in the experiments. We keep increasing the number of multiple task dialogues while reducing the single task dialogues to keep the total number of dialogues constant at $2,150$. The model should be able to learn to generalize to multiple tasks as the set of tasks are the same between the train and test sets with only the nature in which the task is posed by the user is different. We use the Tensor2Tensor \cite{tensor2tensor} framework to run our experiments with \textit(tiny) hyper-parameter setting in the framework. 
\begin{table}[h!t]
\centering
\small
\begin{tabular}{cc|cc}
\toprule
  \multicolumn{2}{c|}{\textbf{Training Data}}  & \multicolumn{2}{c}{\textbf{BLEU}}\\
  Single &  Multiple & Multiple Only & Overall\\
\midrule
$2150$ & $0$ & $7.17$ & $6.81$ \\

$1836$ & $314$ & $7.25$ & $6.87$\\

$1522$ & $628$ & $7.94$ & $7.84$ \\

$1208$ & $942$ & $8.68$ & $8.68$\\

$894$ & $1256$ & $8.83$ & $8.27$ \\

$580$ & $1570$ & $9.33$ & $8.84$ \\

$266$ & $1884$ & $9.10$ & $9.25$ \\

\bottomrule

\end{tabular}
\caption{Ablation study to understand the usefulness of Multiple task dialogues.}
\label{tab:multiple_task_dialog}
\end{table}

As shown in Table \ref{tab:multiple_task_dialog}, the quality of the model improves significantly as number of multiple task dialogues increases. Interestingly, even though the total number of dialogues are kept fixed, the overall validation BLEU score also improves as the number of multiple task dialogues increase in the training set. The results show that the models may be better at decomposing than composing in the domain of goal oriented dialogues or the model at best can only mimic surface level token distribution (Appendix \ref{sec:token-distri}). Though training with more multi-task dialogues can potentially improve the performance, it is not a scalable solution. We will test two of the out-of-the-shelf techniques to improve the task level compositionality in the following section.

\subsection{Zero-shot Compositionality Experiments}

We experiment on Transformer to evaluate the performance on handling zero-shot compositional tasks by training the baseline model only on single task dialogues, and with the proposed data augmentation techniques. The results, in Table \ref{tab:zero_shot}, show that the \emph{Targeted\_Augment} technique increased the performance on multiple-task dialogues by 8.5\% BLEU score while the scores of the model slightly dropped in the performance of all dialogues. 

\begin{table}[!h]
\small
\centering
\setlength\tabcolsep{4pt}
\begin{tabular}{ccc}
\toprule
  \textbf{Data} &  \multicolumn{2}{c}{\textbf{BLEU}}\\
  & Multiple & Overall \\
\midrule  
SNG & 7.17 & 6.81\\
\midrule  
SNG + RS & 7.46 & 7.14 \\
\midrule  
SNG +TS & 7.78 & 7.09 \\

\bottomrule

\end{tabular}
\caption{SNG: Single task dialogues, RS: Random\_Augment Synthetic, and TS: Targeted\_Augment Synthetic. }
\label{tab:zero_shot}
\end{table}

The reason for only a minor BLEU improvement could be due to the noise in generation process. Although the task distributions are matched, the token level distributions appear to be significantly different between the single and multiple-tasks. The results suggest that the method may inject more noise in the token level distribution thereby not improving the model performance significantly.

\subsection{Domain Invariant Transformer}

We compared the proposed architecture and the baseline Transformer model to understand the effects of domain invariant encoder representation towards language generation in multi-task dialogues. We observed from our experiments in Table \ref{tab:multiple-task-final-table} that Domain Invariant Transformer or Transformer model fails to generalize with few-shot multi-task dialogues. The data augmentation techniques too appear to not contribute towards improving the performance. But, Domain Invariant Transformer model improved the performance to a BLEU score when trained only on all of training data, which, though was not the intended objective. Although that seems good, the model is still heavily reliant on human-human multiple domain dialogues and zero-shot or few-shot generalization in compositional dialogues seem quite difficult to achieve.

\begin{table}[h!t]
\centering
\small
\begin{tabular}{p{1.2cm}cc|cc}
\toprule
  \textbf{Model} & \multicolumn{2}{c}{\textbf{Training Data}} & \multicolumn{2}{c}{\textbf{BLEU}}\\
  & Multiple & Synthetic & Multiple & Overall\\
\midrule
\multirow{1}{*}{\parbox{2cm}{Transformer}} & $1.00$ & No & $14.06$ & $14.00$ \\
\midrule
\multirow{2}{*}{\parbox{2cm}{Transformer}} & $0.50$ & Yes & $11.4$ & $12.43$ \\
 & $1.00$ & Yes & $11.89$ & $12.32$ \\
\midrule
\multirow{2}{*}{\parbox{2cm}{Transformer \\ Discriminator}} & $0.50$ & No & $12.24$ & $12.13$ \\
 & $1.00$ & No & $15.06$ & $14.81$ \\
\midrule
\multirow{2}{*}{\parbox{2cm}{Transformer \\ Discriminator}} & $0.50$ & Yes & $11.05$ & $11.60$ \\
 & $1.00$ & Yes & $11.29$ & $12.13$ \\

\bottomrule

\end{tabular}
\caption{0.5 and 1.0 correspond to half and all of multitask samples respectively during training. Synthetic refers to \emph{Targeted\_Augment} dialogues.}
\label{tab:multiple-task-final-table}
\end{table}

The poor performance of the data augmentation techniques can be due to the overwhelming noise in token distribution of input contexts, which skews the language model that the model learns.

\section{Conclusion}

We studied the problem of composing multiple dialogue tasks to predict next utterance in a single multiple-task dialogue. We found that even powerful transformer models do not naturally compose multiple tasks and the performance is severely relied on multiple task dialogues. In this paper, we explored two solutions that only further showed the difficulty of composing multiple dialogue tasks. The challenge in generalizing to zero-shot composition, as observed in the experiments, hints at the possibility of transformer model potentially mimicking only the surface level tokens without understanding the underlying task. The token overlap distribution in Appendix \ref{sec:token-distri} supports the possibility.

\bibliography{main}

\begin{thebibliography}{27}
\expandafter\ifx\csname natexlab\endcsname\relax\def\natexlab#1{#1}\fi

\bibitem[{Baird(1992)}]{aug}
Henry Baird. 1992.
\newblock Document image defect models.
\newblock \emph{Springer}.

\bibitem[{Budzianowski et~al.(2018)Budzianowski, Wen, Tseng, Casanueva, Ultes,
  Ramadan, and Gasic}]{multiwoz}
Pawe{\l} Budzianowski, Tsung-Hsien Wen, Bo-Hsiang Tseng, I{\~n}igo Casanueva,
  Stefan Ultes, Osman Ramadan, and Milica Gasic. 2018.
\newblock Multiwoz - a large-scale multi-domain wizard-of-oz dataset for
  task-oriented dialogue modelling.
\newblock \emph{EMNLP}.

\bibitem[{Byrne et~al.(2019)Byrne, Krishnamoorthi, Sankar, Neelakantan,
  Duckworth, Yavuz, Goodrich, Dubey, Cedilnik, and Kim}]{taskmaster}
Bill Byrne, Karthik Krishnamoorthi, Chinnadhurai Sankar, Arvind Neelakantan,
  Daniel Duckworth, Semih Yavuz, Ben Goodrich, Amit Dubey, Andy Cedilnik, and
  Kyu-Young Kim. 2019.
\newblock Taskmaster-1: Toward a realistic and diverse dialog dataset.
\newblock \emph{EMNLP}.

\bibitem[{Chen et~al.(2016)Chen, Athiwaratkun, Sun, Weinberger, and
  Cardie}]{domain_nlp}
Xilun Chen, Ben Athiwaratkun, Yu~Sun, Kilian~Q. Weinberger, and Claire Cardie.
  2016.
\newblock Adversarial deep averaging networks for cross-lingual sentiment
  classification.
\newblock \emph{TACL}.

\bibitem[{Einolghozati et~al.(2019)Einolghozati, Pasupat, Gupta, Shah, Mohit,
  Lewis, and Zettlemoyer}]{fb}
Arash Einolghozati, Panupong Pasupat, Sonal Gupta, Rushin Shah, Mrinal Mohit,
  Mike Lewis, and Luke Zettlemoyer. 2019.
\newblock Improving semantic parsing for task oriented dialog.
\newblock \emph{arXiv}.

\bibitem[{Eric and Manning(2017)}]{copy-dialog}
Mihail Eric and Christopher Manning. 2017.
\newblock A copy-augmented sequence-to-sequence architecture gives good
  performance on task-oriented dialogue.
\newblock \emph{EACL}.

\bibitem[{Ganin and Lempitsky(2015)}]{domain_1}
Yaroslav Ganin and Victor~S. Lempitsky. 2015.
\newblock Unsupervised domain adaptation by backpropagation.
\newblock \emph{ICML}.

\bibitem[{Kottur et~al.(2017)Kottur, Moura, Lee, and Batra}]{kottur}
Satwik Kottur, Jos{\'e} M.~F. Moura, Stefan Lee, and Dhruv Batra. 2017.
\newblock Natural language does not emerge 'naturally' in multi-agent dialog.
\newblock \emph{EMNLP}.

\bibitem[{Krizhevsky et~al.(2012)Krizhevsky, Sutskever, and Hinton}]{alex}
Alex Krizhevsky, Ilya Sutskever, and Geoffrey~E Hinton. 2012.
\newblock Imagenet classification with deep convolutional neural networks.
\newblock \emph{NeurIPS}.

\bibitem[{Lake and Baroni(2017)}]{lake}
Brenden~M. Lake and Marco Baroni. 2017.
\newblock Still not systematic after all these years: On the compositional
  skills of sequence-to-sequence recurrent networks.
\newblock \emph{Arxiv}.

\bibitem[{Lewis et~al.(2017)Lewis, Yarats, Dauphin, Parikh, and Batra}]{deal}
Mike Lewis, Denis Yarats, Yann~N. Dauphin, Devi Parikh, and Dhruv Batra. 2017.
\newblock Deal or no deal? end-to-end learning for negotiation dialogues.
\newblock \emph{EMNLP}.

\bibitem[{Neelakantan et~al.(2019)Neelakantan, Yavuz, Narang, Prasad, Goodrich,
  Duckworth, Sankar, and Yan}]{neural_assistant}
Arvind Neelakantan, Semih Yavuz, Sharan Narang, Vishaal Prasad, Ben Goodrich,
  Daniel Duckworth, Chinnadhurai Sankar, and Xifeng Yan. 2019.
\newblock Neural assistant: Joint action prediction, response generation, and
  latent knowledge reasoning.
\newblock \emph{Arxiv}.

\bibitem[{Papineni et~al.(2002)Papineni, Roukos, Ward, and
  Zhu}]{papineni2002bleu}
Kishore Papineni, Salim Roukos, Todd Ward, and Wei-Jing Zhu. 2002.
\newblock Bleu: a method for automatic evaluation of machine translation.
\newblock In \emph{Proceedings of the 40th annual meeting on association for
  computational linguistics}. ACL.

\bibitem[{Rojas{-}Barahona et~al.(2017)Rojas{-}Barahona, Gasic, Mrksic, Su,
  Ultes, Wen, Young, and Vandyke}]{pipeline}
Lina~Maria Rojas{-}Barahona, Milica Gasic, Nikola Mrksic, Pei{-}Hao Su, Stefan
  Ultes, Tsung{-}Hsien Wen, Steve~J. Young, and David Vandyke. 2017.
\newblock A network-based end-to-end trainable task-oriented dialogue system.
\newblock \emph{EACL}.

\bibitem[{Schmidhuber(2012)}]{aug_2}
Jurgen Schmidhuber. 2012.
\newblock Multi-column deep neural networks for image classification.
\newblock \emph{CVPR}.

\bibitem[{Sennrich et~al.(2016)Sennrich, Haddow, and Birch}]{back}
Rico Sennrich, Barry Haddow, and Alexandra Birch. 2016.
\newblock Improving neural machine translation models with monolingual data.
\newblock \emph{ACL}.

\bibitem[{Simard et~al.(2003)Simard, Steinkraus, and Platt}]{aug_1}
Patrice~Y. Simard, Dave Steinkraus, and John~C. Platt. 2003.
\newblock Best practices for convolutional neural networks applied to visual
  document analysis.
\newblock \emph{ICDAR}.

\bibitem[{Srivastava et~al.(2014)Srivastava, Hinton, Krizhevsky, Sutskever, and
  Salakhutdinov}]{srivastava14dropout}
Nitish Srivastava, Geoffrey Hinton, Alex Krizhevsky, Ilya Sutskever, and Ruslan
  Salakhutdinov. 2014.
\newblock Dropout: A simple way to prevent neural networks from overfitting.
\newblock \emph{JMLR}.

\bibitem[{Sun et~al.(2018)Sun, Yeh, Hwang, Ostendorf, and Xie}]{speech_domain}
Sining Sun, Ching{-}Feng Yeh, Mei{-}Yuh Hwang, Mari Ostendorf, and Lei Xie.
  2018.
\newblock Domain adversarial training for accented speech recognition.
\newblock \emph{Arxiv}.

\bibitem[{Tzeng et~al.(2015)Tzeng, Hoffman, Darrell, and Saenko}]{domain_2}
Eric Tzeng, Judy Hoffman, Trevor Darrell, and Kate Saenko. 2015.
\newblock Simultaneous deep transfer across domains and tasks.
\newblock \emph{ICCV}.

\bibitem[{Vaswani et~al.(2018)Vaswani, Bengio, Brevdo, Chollet, Gomez, Gouws,
  Jones, Kaiser, Kalchbrenner, Parmar, Sepassi, Shazeer, and
  Uszkoreit}]{tensor2tensor}
Ashish Vaswani, Samy Bengio, Eugene Brevdo, Francois Chollet, Aidan~N. Gomez,
  Stephan Gouws, Llion Jones, \L{}ukasz Kaiser, Nal Kalchbrenner, Niki Parmar,
  Ryan Sepassi, Noam Shazeer, and Jakob Uszkoreit. 2018.
\newblock Tensor2tensor for neural machine translation.
\newblock \emph{Arxiv}.

\bibitem[{Vaswani et~al.(2017)Vaswani, Shazeer, Parmar, Uszkoreit, Jones,
  Gomez, Kaiser, and Polosukhin}]{transformer}
Ashish Vaswani, Noam Shazeer, Niki Parmar, Jakob Uszkoreit, Llion Jones,
  Aidan~N Gomez, \L~ukasz Kaiser, and Illia Polosukhin. 2017.
\newblock Attention is all you need.
\newblock \emph{NeurIPS}.

\bibitem[{Vinyals and Le(2015)}]{conv-seq2seq}
Oriol Vinyals and Quoc~V. Le. 2015.
\newblock A neural conversational model.
\newblock \emph{CoRR}.

\bibitem[{Wen et~al.(2015)Wen, Gasic, Mrk{\v{s}}i{\'c}, Su, Vandyke, and
  Young}]{sclstm}
Tsung-Hsien Wen, Milica Gasic, Nikola Mrk{\v{s}}i{\'c}, Pei-Hao Su, David
  Vandyke, and Steve Young. 2015.
\newblock Semantically conditioned lstm-based natural language generation for
  spoken dialogue systems.
\newblock \emph{EMNLP}.

\bibitem[{Wen et~al.(2017)Wen, Miao, Blunsom, and Young}]{latent}
Tsung{-}Hsien Wen, Yishu Miao, Phil Blunsom, and Steve~J. Young. 2017.
\newblock Latent intention dialogue models.
\newblock \emph{{ICML}}.

\bibitem[{Xu et~al.(2017)Xu, Zhang, Zhang, Lin, and Wang}]{video}
Gao Xu, Yongming Zhang, Qixing Zhang, Gaohua Lin, and Jinjun Wang. 2017.
\newblock Domain adaptation from synthesis to reality in single-model detector
  for video smoke detection.
\newblock \emph{ArXiv}.

\bibitem[{Xu and Yang(2017)}]{domain_nlp_1}
Ruochen Xu and Yiming Yang. 2017.
\newblock Cross-lingual distillation for text classification.
\newblock \emph{ACL}.

\end{thebibliography}
\bibliographystyle{acl_natbib}    
\newpage
\ 
\newpage
\appendix
\section{Preprocessing}
\label{preprocessing}

\begin{figure*}[ht]
\centering
\tiny
\begin{minipage}{5 cm}
\vspace{1em}
\textcolor{purple}{\textbf{USER:}I'm looking for the Lensfield Hotel. Can you help me find it ?}\par
\textcolor{purple}{\textbf{ASSISTANT:}Yes of course, it is located in 53-57 lensfield road. Would you like me to book you a reservation ?}\par
\textcolor{purple}{\textbf{USER:} No thank you. Could you give me the postcode and tell me if they have free parking ?}\par
\textcolor{purple}{\textbf{ASSISTANT:} Yes, they have free parking. The postcode is cb21en. Is there anything else you'd like to know about this hotel ?}\par
\textcolor{purple}{\textbf{USER:}No, that's all I needed. Thanks! Bye!}\par
\textcolor{purple}{\textbf{ASSISTANT:} You are welcome. Goddbye!}\par
\end{minipage}
\hspace{0.2cm}
\begin{minipage}{5 cm}
\vspace{1em}
\textcolor{teal}{\textbf{USER:}I'd like to find a train that leaves after 17:00 on Saturday.}\par
\textcolor{teal}{\textbf{ASSISTANT:} Absolutely, I need departure and destination information and a time you need to arrive by.}\par
\textcolor{teal}{\textbf{USER:} I will be departing from Broxbourne and I'd like to go to Cambridge.}\par
\textcolor{teal}{\textbf{ASSISTANT:} I have 7 available. The first leaves at 17:32, and arrives at 18:32. Would you like me to book that for you ?}\par
\textcolor{teal}{\textbf{USER:}I am not looking to book at the moment, can I just get the train ID? Thanks.}\par
\textcolor{teal}{\textbf{ASSISTANT:} Of course! TR0811 leaves at 17:32, arrives at 18:32, with a duration of 60 minutes. The price is 14.32 pounds. May I assist you with anything else ?}\par
\textcolor{teal}{\textbf{USER:} That is all I need for today. Thank you. Good bye.}\par
\textcolor{teal}{\textbf{ASSISTANT:} Okay glad I could help you. Enjoy your trip.}\par
\end{minipage}
\hspace{0.2cm}
\begin{minipage}{5 cm}
\vspace{1em}
\textcolor{purple}{\textbf{USER:}I'm looking for the Lensfield Hotel. Can you help me find it ?}\par
\textcolor{purple}{\textbf{ASSISTANT:}Yes of course, it is located in 53-57 lensfield road. Would you like me to book you a reservation ?}\par
\textcolor{purple}{\textbf{USER:} No thank you. Coul you give me the postcode and tell me if they have free parking ?}\par
\textcolor{purple}{\textbf{ASSISTANT:} Yes, they have free parking. The postcode is cb21en. Is there anything else you'd like to know about this hotel ?}\par
\textcolor{teal}{\textbf{USER:}I'd like to find a train that leaves after 17:00 on Saturday.}\par
\textcolor{teal}{\textbf{ASSISTANT:} Absolutely, I need departure and destination information and a time you need to arrive by.}\par
\textcolor{teal}{\textbf{USER:} I will be departing from Broxbourne and I'd like to go to Cambridge.}\par
\textcolor{teal}{\textbf{ASSISTANT:} I have 7 available. The first leaves at 17:32, and arrives at 18:32. Would you like me to book that for you ?}\par
\textcolor{teal}{\textbf{USER:}I am not looking to book at the moment, can I just get the train ID? Thanks.}\par
\textcolor{teal}{\textbf{ASSISTANT:} Of course! TR0811 leaves at 17:32, arrives at 18:32, with a duration of 60 minutes. The price is 14.32 pounds. May I assist you with anything else ?}\par
\textcolor{teal}{\textbf{USER:} That is all I need for today. Thank you. Good bye.}\par
\textcolor{teal}{\textbf{ASSISTANT:} Okay glad I could help you. Enjoy your trip.}\par
\end{minipage}
\caption{An example of combining two single-task dialogues in \textcolor{purple}{color1} and \textcolor{teal}{color2} together to form a single multi-task dialogue.}
\label{example1}
\end{figure*}

The MultiWoZ 2.0 dataset has a JSON metadata that maintains a dictionary of slot-value pairs provided by the user to the agent in every utterance. We use this metadata to construct a local and a global knowledge of slot-value shared by the user and split to relabel the dataset for single domain and multidomain dialogues. The preprocessing step removed the noise in the labeling of dialogues. We used this approach to keep a test set of multi-domain dialogues to evaluate the model performance on compositional tasks. On the clean split of single domain dialogues we generate synthetic multidomain dialogues using two different approaches:

\subsection{Random Synthetic (RS)}

In this approach, we pick a single task dialogue $^iD^{SNG}$ and randomly select a set of \emph{K} single task dialogues,$\left(^iD^{SNG}_{noise}\right)_{k=1}^K$, to inject noise in $D^{SNG}$. With an hyperparameter, \emph{percentCopy}, we select the number of utterances to be copied from every dialogue in the set noiseDialogues and add it as a prefix to $D^{SNG}$. This results in $K$ negative samples of synthetic multidomain dialogues, $\left(^iD^{MUL}_{RS}\right)_{k=1}^K$, for every single domain dialogues in the dataset.

\subsection{Targetted Synthetic (TS)}

We bucket the single domain dialogues based on the conversation domain (\emph{taxi, hotel, attraction} etc.,). Similarly, we bucket the multi-task dialogues in the training set to measure the topic distributions in multi-task dialogues. Using the computed distribution of composite tasks in \emph{true} multidomain dialogues and the domain label of every $^iD^{SNG}$, we constrain the selection of random dialogues to conform to the training distribution of \emph{true} composite tasks in the training set. The hyperparameters and the remainder of the procedure is similar to RS except when combining the single domain dialogues from two different domains $\left(^iDom,^jDom\right)$, we inject the topic change exchanges randomly sampled from $TC^{\left(^jDom1,^iDom2\right)}$.

For training the proposed Domain Invariant Transformer model, we create the labels for the auxiliary tasks using the preprocessing steps used to split the dataset into single and multi-domain dialogues

\subsection{Experiments varying $\alpha$}

\begin{table}[H]
\centering
\small
\begin{tabular}{ccc}
\toprule
  \textbf{$\alpha$} & \textbf{BLEU (MUL)}&\textbf{BLEU}(BOTH)\\
\midrule
0.0 & 14.07 & 13.94\\
\midrule
0.00001 & 13.74 & 13.31\\
\midrule
0.0001 & 14.13 & 14.11\\
\midrule
0.001 & 15.06 & {\bf 14.81}\\
\midrule
0.01 & 14.61 & 14.40\\
\midrule
0.1 & 14.70 & 14.41\\
\bottomrule

\end{tabular}
\caption{Varying the $\alpha$ to understand the effect of the discriminator on decoder performance}
\label{tab:alpha}
\end{table}

We experimented with different values of $\alpha$ to understand the influence of the discriminator loss. The results in Table \ref{tab:alpha} show that Domain Invariant Transformer performed better when $\alpha$ is $0.001$. The experiment also shows consistent performance improvement in all the experiments with different $\alpha$ highlighting the usefulness of training an auxiliary network to train domain invariant encoder representations.

\section{Token distribution}
\label{sec:token-distri}
\begin{table}[h]
    \centering
    \small
    \subfigure[Table 1]{
    \begin{tabular}{c|c}
    \midrule
         \textbf{MUL Train} &  492688\\
         \textbf{SNG Valid} &  16907\\
         \textbf{Intersection} & 9238 \\
         \textbf{\% Unseen} & 45.36\\
    \end{tabular}}
    \subfigure[Table 2]{
    \begin{tabular}{c|c}
        \midrule
        \textbf{MUL Train} & 492688 \\
         \textbf{MUL Valid} &  104261\\
         \textbf{Intersection} & 48076 \\
         \textbf{\% Unseen} & 53.89\%\\
    \end{tabular}}
    \\
    \subfigure[Table 3]{
    \begin{tabular}{c|c}
        \midrule
        \textbf{SNG Train} & 124038 \\
         \textbf{MUL Valid} &  104261\\
         \textbf{Intersection} & 22254 \\
         \textbf{\% Unseen} & 78.66\%\\
    \end{tabular}}
    \subfigure[Table 4]{
    \begin{tabular}{c|c}
        \midrule
        \textbf{SNG Train} & 124038 \\
         \textbf{SNG Valid} &  16907\\
         \textbf{Intersection} & 6562 \\
         \textbf{\% Unseen} & 61.19\%\\
    \end{tabular}}
    \\
    \subfigure[Table 5]{
    \begin{tabular}{c|c}
        \midrule
         \textbf{SNG+MUL Train} &  568674\\
         \textbf{SNG Valid} &  104261\\
         \textbf{Intersection} & 49999 \\
         \textbf{\% Unseen} & 52.04\%\\
    \end{tabular}}
    \subfigure[Table 6]{
    \begin{tabular}{c|c}
        \midrule
         \textbf{SNG+MUL Train} &  568674\\
         \textbf{SNG Valid} &  16907\\
         \textbf{Intersection} & 9746 \\
         \textbf{\% Unseen} & 42.36\%\\
    \end{tabular}}
    \caption{Analysis of 4-gram overlap across different combinations of train and validation splits that were used in the experiments. The analysis show that the \%Unseen in validation set is higher when training with SNG (Single domain dialogues) but considerably lower when trained with MUL. The composition task requires models to understand the underlying task structure but the data distribution and performance of transformer strongly correlate to show that the transformer model at best mimics the surface level token distribution than understanding the nature of task.}
    \label{tab:token-distibution}
\end{table}
We analyze the token distribution in the dataset to understand the negative result further. We observed that despite the task distributions are matched the underlying token distribution in different set up is not (Table \ref{tab:token-distibution}). We looked at the overlap of the distribution of 4-grams in conversations on the different splits we used for training. We observed that Multi-task dialogues (MUL) training set has as much 4-gram overlap with MUL Valid and SNG (Single task dialogues) Valid sets as the combined (SNG + MUL) training data.

The analysis raises doubts in the performance of transformer model with increased MUL train dialogues that the performance improvement cannot be only because of the model's ability to decompose multiple tasks but may be because the MUL train has higher 4-gram overlap with SNG Valid and MUL Valid. This shows that despite the dialogues carrying rich information in task oriented dialogues, the model at best only mimics the surface level token distribution. Hence, it is not clear if the Transformer model can generalize to multi-task dialogues with an understanding of the underlying task structure. 

\end{document}